# Large Language Models Powered Multiagent Ensemble for Mitigating Hallucination and Efficient Atrial Fibrillation Annotation of ECG Reports

Jingwei Huang[1], Kuroush Nezafati[1], Ismael Villanueva-Miranda[1], Zifan Gu[1], Yueshuang Xu[1], Ann Marie Navar[1,2], Tingyi Wanyan[1], Qin Zhou[1], Bo Yao[1], Ruichen Rong[1], Xiaowei Zhan[1], Guanghua Xiao[1], Eric D. Peterson[1,2], Donghan M. Yang[1*], Wenqi Shi[1*], Yang Xie[1*]

1. Quantitative Biomedical Research Center, Peter O'Donnell School of Public Health, University of Texas Southwestern Medical Center, 5323 Harry Hines Blvd, Dallas, TX, USA 75390

2. Department of Internal Medicine, University of Texas Southwestern Medical Center, 5323 Harry Hines Blvd, Dallas, TX, USA 75390

**\*Correspondence to**:

Wenqi Shi. Email: Wenqi.Shi@UTSouthwestern.edu

Donghan M. Yang. Email: Donghan.Yang@UTSouthwestern.edu

Yang Xie. Email: Yang.Xie@utsouthwestern.edu

# Abstract

**Objective**

Atrial fibrillation, the most common sustained cardiac arrhythmia, significantly increases the risks of thromboembolism, stroke, and heart failure. Its asymptomatic and intermittent nature often delays diagnosis until serious cardiovascular events occur. To address this issue, we aim to automatically generate atrial fibrillation and flutter (AF) labels from electrocardiogram (ECG) reports, enabling large-scale data annotation for downstream data-driven applications in AF detection and risk prediction.

**Materials and Methods**

In this study, we analyzed 623,566 ECG reports from the MIMIC-IV dataset. We developed a multi-agent ensemble method using Large Language Models (LLMs) to automate digital phenotyping of AF from free-text ECG reports, incorporating uncertainty handling mechanisms to reflect diagnostic ambiguity, and mitigating hallucinations.

**Results**

Our ensemble method achieved 98% accuracy on a manually labeled ECG test set, significantly outperforming individual LLMs and reducing hallucination errors. We further demonstrated its generalizability by identifying social determinants of health from clinical notes with 99.7% accuracy.

**Discussion**

Manual labeling of large-scale healthcare data remains labor-intensive, costly, error-prone, and requires specialized domain expertise, limiting machine learning applications in clinical research. The proposed method provides an effective and efficient solution, substantially reducing human effort (from ~6.8 person-years to 2 weeks of computation) while maintaining high accuracy.

**Conclusion**

Our study demonstrates that the LLM multiagent ensemble approach effectively automates large-scale EHR data annotation and successfully generalizes to diverse clinical annotation tasks, as illustrated by identifying social determinants of health. Additionally, our ensemble approach outperforms individual LLMs and significantly reduced hallucination errors, highlighting its potential for reliable and scalable deployment in healthcare data annotation.

# Introduction

Atrial fibrillation (AFib) is the most common sustained arrhythmia and contributes significantly to major adverse cardiovascular diseases such as thromboembolism, stroke, and heart failure.[1-3] Its asymptomatic and paroxysmal nature often delays diagnosis until severe outcomes occur.[4 5] Timely diagnosis and accurate identification of AFib are thus essential for effective clinical management and prevention of complications.

The standard 12-lead electrocardiogram (ECG) is a non-invasive, cost-effective tool for cardiologists to assess cardiac function and detect AFib.[6] An ECG captures cardiac electrical activity, allowing specialists to identify pathological patterns indicative of AFib, including irregular R-R intervals, absence of repetitive P-waves, and chaotic atrial activity.[7] However, interpretation of ECG reports is complicated by mixed rhythms, artifacts, and clinical uncertainty, often reflected through qualifiers such as "possible" or "probable" as suggested by established guidelines,[8-10] Consequently, manual review and annotation of ECG reports for AFib is labor-intensive, time-consuming, expensive, and prone to inconsistencies, creating a critical barrier to utilizing large ECG datasets in artificial intelligence and machine learning research for AF diagnosis and risk prediction.

Recent advances in Artificial Intelligence (AI), particularly deep learning, have demonstrated strong potential for automated and scalable analysis of ECG signals to detect AFib.[4 11 12] Large-scale ECG datasets, such as the Medical Information Mart for Intensive Care (MIMIC)-IV ECG dataset[13 14] containing 623,566 ECG studies, present a valuable data resource for AI-driven cardiovascular research. Nonetheless, a major bottleneck remains: the lack of annotated ECG data required for supervised deep learning applications. An ECG report captures generic and elementary cardiac features and suggests cardiac abnormalities in free text, but it does not directly provide a structured conclusion about whether AFib presents. Thus, automating the labeling of AFib from clinical ECG reports becomes essential to enable large-scale research and clinical applications.

Large language models (LLMs) [15-21] have recently emerged as a powerful solution for processing unstructured clinical texts and automating annotation tasks. Trained on massive amounts of text data, LLMs are effective for tasks such as classification, summarization, and information extraction, substantially reducing the manual workload associated with data labeling. Yet, individual LLMs vary considerably in performance and may introduce labeling errors or hallucinations, limiting their reliability for clinical decision-making. Furthermore, directly using LLMs for labeling tasks without a structured process (i.e., agentic workflow) may overlook essential aspects of interpretability, quality control, and error correction.

Among the issues, hallucination is a critical challenge and has received much attention.[22-26] It is particularly concerned with LLMs' medical applications due to the consequences of false hallucinated content. Word "hallucination" originated from the Latin "hallucinari" or the Greek

"alucinari", meaning to wander in the mind; as a psychological concept, hallucinations was defined as the unreal perceptions that occur in the absence of corresponding external stimulus[27 28]. In Natural Language Generation (NLG), "hallucination" is borrowed to refer to the generated content that is "nonsensical" or "unfaithful" to the source content[23]. Hallucination is associated with two properties: (1) faithfulness, which is the extent of the generated content being consistent and truthful to the source content, thus being an antonym to "hallucination"; (2) factuality is the quality of being actual or fact grounded. Based on the range of facts, hallucinations are categorized in two types: intrinsic and extrinsic. Intrinsic hallucinations are the generated content inconsistent with the source content. Extrinsic hallucinations are the generated content that cannot be verified with the input source content but may be verifiable with the extended "world knowledge"[22 23]. The world knowledge is a hypothesis of the universal knowledge that the LLMs acquired through training with numerus various text data. In the real world, each individual LLM has its strength and limitations or bias in knowledge due to different training data, different training strategies, and different training processes. Another more straightforward taxonomy of hallucinations[24] covers: (1) input-conflicting hallucination, which is inconsistent with the input, including instruction and task input; (2) context-conflict hallucination, which is inconsistent with the generated content, or self-contradiction; (3) fact-conflict hallucination, which is inconsistent with the established world knowledge. Many hallucination mitigation strategies have been proposed[24 26]. This article presents an LLM multiagent ensemble method to mitigate hallucinations.

To overcome those challenges, we investigate a LLMs powered multiagent ensemble approach for automatic data labeling of a large volume of ECG reports. The convergence of LLMs, intelligent agents and multiagent systems, as well as ensemble learning is advancing LLM-based agents in healthcare.[29-31] Qiu[29] presents an inspiring vision to wrap LLM in an agentic workflow within an agent for reliable and functionally enhanced medical applications. Wang[31] provided a survey on LLM agents in medicine. Yang[30] applied ensemble LLMs for medical QA tasks and evaluated with 3 public datasets. We apply LLMs multiagent ensemble to solve a real-world data labeling problem – AF annotation for MIMIC-IV ECG data. Our approach uses each LLM as reasoning engine of each agent representing an expert with specific expertise and then combines outputs from different LLM agents to produce more accurate and reliable labels with significant reduced hallucinations. **Fig. 1** illustrates the overall workflow of our LLM multiagent ensemble approach for ECG data labeling. This approach enables leveraging LLMs' extensive common knowledge with controlled, transparent, and manageable applications, rather than using LLMs directly as end-to-end tools. We applied this LLM multiagent ensemble method to automate labeling of 623,566 ECG reports from the MIMIC-IV dataset,[13 14] aiming for high accuracy, improved efficiency, and substantial reduction in hallucination errors. Additionally, we evaluated the generalizability of this method by applying it to a separate clinical annotation task—identifying social determinants of health (SDOH, such as socioeconomic status, education, employment, social support, and health behaviors like smoking and alcohol use[32-34]) from the social history sections of 1,405 clinical notes in MIMIC-SDBH dataset[35]— further highlighting its potential for broad clinical use. In both applications, our ensemble method achieved high accuracy while significantly reducing the time and effort required for manual annotation. Our research addresses a fundamental challenge in healthcare AI, the efficient creation of high-quality labeled datasets, while providing specific value for cardiovascular research through the comprehensive labeling of a major ECG database.

# Methods

## Problem formulation and prompt design

According to AHA/ACC guideline [3 10], atrial fibrillation (AFib) and atrial flutter frequently coexist; and they share similar thromboembolic risk profiles[36-39]. Therefore, some research [4 40] "group atrial flutter with AF (atrial fibrillation)". For our downstream machine learning research about AF and their thromboembolic risks, such as stroke,[41] we define AF category as "AFib or AFL" in the automatic data labeling task.

Generally, uncertainty broadly exists in medicine. Clinical practice uses various forms to express uncertainties, such as "Cannot be determined", "Not specified". As addressed in Introduction, mixed rhythms, artifacts, and clinical uncertainty make diagnosis not absolute. AHA/CCC/ACCF guideline[8] represents uncertainty with modifiers as suggested in the table 3 of the recommendation[8] and incorporate probability to diagnosis for "Computerized Interpretation of the ECG"(p.1318)[9]. To reflect this clinical diagnosis uncertainty in the automatic annotation process with LLM multiagent, in addition to ECG report categories of AF and Non-AF, we also include an "Uncertain" category that has the semantics of "AF or Non-AF" and is used for informing users that further human review needed, due to uncertain factors in those ECG reports.

Our primary objective is to extract the cardiologist's assessments from ECG reports, rather than independently diagnosing AFib based on raw signal data. We developed specialized prompts to guide LLMs in identifying these clinical opinions and their supporting evidence within the free-text reports. The prompts were iteratively refined using a development subset of 200 reports (AF200) randomly selected from the MIMIC-IV dataset (Fig. 2).

In prompt design, we follow a principle to use categories and terms as close to the language used in input data (here, ECG reports) as possible, thus reducing the difficulty of the tasks to LLMs and allowing LLMs achieve better performance. With this in mind, we request LLMs to make classification in five categories ("AF," "Probable AF," "Possible AF," "Non-AF," "Not Specified"). For LLMs' output, we conduct postprocessing that merges "Probable AF" into AF as they're almost certain to be AF, and merges "Possible AF" (cases where AF is one of two or more possibilities) and "Not Specified" (no information) as "Uncertain" category that needs human review. We also query LLMs to provide "AF_pr" (probability of AF), which we did not use in voting, as LLMs do not always give the score consistently.

Generally, prompt design process mathematically is a multi-attribute decision making process. Several metrics, such as false negative rate for AF cases, false positive rate, and overall F1 score, need to be considered. In our prompt engineering process, we comprehensively consider the numbers of false predictions in different positions of the confusion matrix, where different false predictions have different levels of priority to reduce, as shown in Fig. 4a. In the prompt engineering process, we found that a significant number of errors are due to uncertainty factors, (such as uncertainty modifiers, noises, multiple rhythms, ambiguous description, or no sufficient information,) which a LLM classified differently and needs more instructions or more background knowledge. To address the issues, we tactically introduced "Uncertain" category to indicate the present of high uncertainty, and we also add some necessary background knowledge in the system message.

## Data pre-processing and filtering

The MIMIC-IV ECG dataset contains 623,566 unlabeled ECG reports documenting cardiac electrical activity across diverse patient populations. To optimize computational efficiency, we implemented a preliminary screening process using logical filters[42 43] based on domain knowledge. AHA/CCC/ACCF guideline[8] has standardized "Primary Statements", covering 117 primary diagnostic statements under 14 categories, including "Atrial fibrillation" and "Atrial flutter". Our exploratory analysis further confirmed that all AF cases indeed contained AF-related terminology — following the guideline, regularly cardiologists make primary diagnostic statements (even though in free text), explicitly using atrial fibrillation or flutter terms for diagnosed AF cases in the ECG reports. This enables us to develop logical filters using regular expressions to isolate AF-relevant reports from the dataset. The limitation of this assumption is that the method may miss AF cases, if some individual cardiologists do not follow the guideline by not stating the standardized primary statements. These regular expressions account for term variations, abbreviations, typographical errors, and multilingual representations (e.g., "fibri.," "fibrillacion," "fabrillation"). The logical filter for "fibrillation" and its variants can be defined with a regular expression as follows:

$$r'\backslash b((([fF][ia]br[ia]l\{1,2\}a[tcs]i[oó]n)|([fF]ibri\backslash.*)|(A[fF]ib)|(AF))\backslash b'$$

As illustrated in **Fig. 2**, this filtering process identified 68,165 AF-relevant cases (10.93% of the total dataset), while 555,401 cases were classified as AF-irrelevant. The medical experts in our team manually verified 200 randomly selected AF-irrelevant cases, confirming the absence of any AF diagnoses. This approach allowed us to focus computational resources on the AF-relevant subset, which required more nuanced interpretation as these reports might mention AF terms in various contexts (e.g., comparing with previous studies, noting secondary rhythms, or documenting suspected AF with uncertainty qualifiers). Under supervision of two senior cardiology experts and board-certified cardiologists, a board-certified internist with expertise on ECG interpretation manually labeled 200 cases (named as dataset AF200) and 600 additional cases for testing performance (named as AF600 or test dataset), randomly selected from the AF-relevant subsets. AF200 and have similar distribution. AF200 has 87.9% of AF cases, 7.1% of Non-AF cases, and 5.1% of Uncertain cases; AF600 has 86.7% of AF, 8.7% of Non-AF, and 4.7% of Uncertain cases. We randomly selected 60 cases (20 from AF200; 20 from NAF200, 20 from AF600) for peer validation. Another MD, also guided by senior cardiology experts, annotated the 60 cases independently, achieved 100% consistency with the first labeling. We used the AF200 subset for prompt engineering to develop the prompt for labeling. With this prompt, the LLM multiagent ensemble method was applied to the rest of AF-relevant cases, with performance evaluated against the expert-labeled test set.

We conducted verification of the logical filtering in two means: (1) Independently implemented two alternative filters, one is simple and straightforward,

$$r'\backslash b(fibrillation|flutter|AFib|AFL|AF)\backslash b'$$

and the second is more complex and consider broader range of possible expressions.

$$r'\backslash b([fF][ia]br[ia]lla[tcs]ion|[fF]ibri\backslash.*|[fF]lutter|A[fF]ib|AFL|AF)\backslash b'$$

The difference of filtering out is only 2 cases (which spell fibrillation as "fibrilation") among 555,401 AF-irrelevant cases, i.e. at rate of 0.00036%. These two filters got the same result. (2) In our manual verification of ground truth annotation by a different annotator, we included 20 cases from 200 AF-irrelevant cases, and the result remains the same.

## LLM selection and multiagent architecture

To identify most suitable LLMs for ECG report interpretation, we systematically evaluated a diverse pool of LLMs from different architectural "families" with different parameter size, pre-trained or finetuned with different training data by different providers. The tested LLMs cover GPT variants, Llama3, gemma models, fine-tuned Llama2-based models, and mixture-of-experts (MoE) models.[44 45] Considering computational efficiency, accuracy, and cost, with a human-directed heuristic search for minimizing the number of false positive (FP) and false negative (FN) prediction of AF, we selected 7 open source LLMs with acceptable performance on the ECG data, including beluga (70b),[46] gemma (7b),[47] llama3 (70b-instruct),[48] mistral-openorca (7b)[49], openhermes (7b),[50] qwen (72b),[51] and qwen2 (72b).[52] As an example, we did not select GPT models, although GPT-4o achieved high accuracy, but it also has high cost for a large dataset; GPT3.5-Turbo is cheaper but its FN up to 20. Each LLM was tested with a standardized prompt (Fig. 3) to classify ECG reports into "AF," "Non-AF," or "Uncertain". Fig. S1 presents the performance of each LLM.

## Ensemble voting and label aggregation

We conceptualized each LLM agent's label prediction as a vote within a decision-making committee. The final classification of each ECG report was determined through a structured voting mechanism that aggregates these individual predictions. The rule of highest vote with a minimal threshold is applied. Formally, let

$$V_0 = \{AF, NonAF, Uncerrtain, NA, ...\} \quad (1)$$

be the set of labels that LLMs may generate. $V$ is the set of valid votes.

$$V = \{AF, NonAF, Uncerrtain\} \quad (2)$$

Let $v_i \in V_0$ be the vote of agent $i$, $i = 1, 2, ..., n$.

Define a membership function to facilitate counting votes on a valid label as follows.

$$f(x, v) = \begin{cases} 1 & v = x \land x \in V; \\ 0 & \text{otherwise.} \end{cases} \quad (3)$$

which means if vote $v$ is label $x$ in $V$, $f(x, v)$ is 1, otherwise 0.

Define votes counting function as follows.

$$votes(x) = \sum_{i=1}^{n} f(x, v_i) \quad (4)$$

Then, the set of winners (which possibly tie on votes) is

$$Y = arg_{x \in V} \ max \ votes(x) \quad (5)$$

For simplicity and generality, we adopt simple and most common ensemble labeling rules as follows.

**Ensemble rule (highest vote with winning threshold):**

$$w = \begin{cases} y & y \in Y \land |Y| = 1 \land \frac{votes(y)}{\sum_{x \in V} votes(x)} > \theta; \\ Review & \text{otherwise.} \end{cases} \quad (6)$$

Generally, the threshold $\theta$ is problem-specific and depends on the number of possible labels and the trade-off between performance and the cost for human review. For example, using a majority voting rule with a threshold over 50%, the winning label must receive at least half of the votes (at least 4 out of 7 votes in our application). If no label meets this threshold, or if a voting tie occurs, the case is marked as "Review" for further evaluation by a human expert. Alternatively, when using a simple highest vote rule (set $\theta$ as 0), the label receiving the highest number of votes is selected regardless of the majority. For example, if 3 agents vote "AF," 2 agents vote "Uncertain," and 2 agents vote "Non-AF," the highest vote rule selects "AF" as the winning label, whereas the majority rule with a 50% threshold labels the same case as "Review.".

## Implementation details

The LLM multiagent ensemble architecture is configured through a YAML file that specifies: (1) the LLMs to use, (2) the framework for launching each LLM, (3) input data paths, (4) prompt file paths, (5) expected output JSON structure, and (6) additional execution parameters. Supplementary **Fig. S6** details the architecture, workflow, and data flow of our implementation. Based on this configuration, the system constructs an LLM multiagent ensemble application object that creates and manages a group of specified LLM agents. Each agent uses the designated framework to load and run its assigned language model. Currently supported frameworks include: (a) Ollama; (b) Transformers with HuggingFace hub for accessing open-source LLMs; (c) OpenAI API and (d) Azure OpenAI to access GPT model family. The ensemble application first creates a configuration file for each individual LLM agent, then uses these configurations to instantiate the agents. Each agent receives identical prompts and input data, producing JSON-formatted outputs in its assigned directory. The agents include functionality to correct common JSON syntax errors, extract structured data, and output results in CSV format. The ensemble application then aggregates these CSV files, applies the configured voting rules, calculates the final ensemble result, and evaluates performance metrics. For the ECG labeling task, we prepared a specialized prompt (Supplementary Fig. S1) and applied our ensemble methodology to all AF-relevant cases. A similar workflow was implemented for the SDOH identification task, with appropriate modifications to the prompt design and input data format. The complete source code for our implementation will be made available on GitHub.

## Results

### LLM multiagent ensemble labeling of the ECG dataset

Using the proposed LLM multiagent ensemble method, we labelled whole MIMIC-IV ECG dataset of 623,566 ECG reports. **Fig. S3** illustrates the final label distribution over the dataset. As shown in **Fig. S3a**, the complete ECG dataset consists of two subsets: (1) 555,401 AF-irrelevant cases, which were labeled as Non-AF based on our filtering procedure, and (2) 68,165 AF-relevant cases, which were labeled using the LLM multiagent ensemble method. The labels include: "AF" (either "atrial fibrillation" or "atrial flutter"), "Non-AF", "Uncertain" (no sufficient information to tell it is AF or non-AF), and "Review" (high disagreement among the voting of LLM agents, thus need human experts to review). In practice, both labels of "Uncertain" and "Review" are the cases that need review. The distribution rate of each label over the whole ECG dataset is provided in the right panel.

### Performance Evaluation on Test Dataset

**Fig. 4** shows the performance of the labeling of a randomly selected and labeled 600 cases as test dataset. **Fig. 4a** displays the confusion matrix for the most stringent ensemble configuration, requiring unanimous consensus (100% agreement) among all LLM agents. At full consensus (7/7 votes), the method achieved 99% accuracy (F1: 99.1%), though 80 cases required human review. The confusion matrices in **Figs. 4b** and **4c** illustrate performance with more permissive voting thresholds of 4/7 (≥57%) and 3/7 (≥43%), respectively. At a more practical 50% consensus threshold (4/7 votes), both accuracy and F1 was 98.2%, with only one case flagged for review. Lowering the threshold further (≤3 votes) slightly decreased accuracy and F1 to 98.0% and 98.1%, but completely eliminated the need for human review.

To reflect the clinical reality that different classification errors carry varying levels of importance, we highlighted three distinct levels of misclassification severity in the confusion matrices, as illustrated in **Fig. 4a**. **Fig. 4d** summarizes key performance metrics across different voting thresholds, including recall and specificity for AF detection. This analysis reveals a clear trade-off: as the winning threshold increases, accuracy improves at the cost of a substantially higher review rate, particularly at thresholds approaching 100%. Generally, increasing the consensus threshold improved labeling accuracy but significantly increased the human review burden. These results demonstrate that our LLM multiagent ensemble method can effectively automate ECG report labeling at scale while maintaining high accuracy and providing a configurable mechanism to balance automation efficiency against the need for expert verification in ambiguous cases.

## Application to SDOH Identification

To demonstrate the generalizability of our approach, we applied the LLM multiagent ensemble method to identify social determinants of health by using a randomly selected subset of 1405 notes (20%) from a curated MIMIC-III dataset[35] on Social Behavioral Determinants of Health. We selected two variables employment status (reflecting economic status) and housing status (reflecting living environment status) for this purpose. Ground-truth labels were previously established for both variables, with an additional "Uncertain" category for indeterminate cases.

Results for employment status identification are presented in **Fig. 5 (a to e)**. Using the simple highest vote rule without a minimum threshold (0/7), our ensemble method achieved 95% overall accuracy and a 95.4% Jaccard score for identifying adverse employment conditions, with only 0.4% of cases (2 instances) requiring review. Performance remained consistent across threshold values up to 50% (4/7). When implementing more stringent thresholds, both accuracy and the Jaccard score for "Adverse" classifications improved—reaching 97.6% accuracy and 96.2% Jaccard at the unanimous threshold (7/7)—at the cost of a substantially increased review rate of 39.4%.

**Fig. 5 (f to j)** presents results for housing status identification, where our method demonstrated even stronger performance. Using highest vote rule, we achieved 99.7% accuracy and an 81.2% Jaccard score with zero cases requiring review. Increasing the threshold to 5/7 (71.4%) further improved accuracy to 99.9% and the Jaccard score to 92.9%, with only a modest increase in review rate to 1.5% (13 cases out of 892 with determinate ground truth).

## Performance comparison with individual LLMs

We conducted comprehensive comparisons between our ensemble approach and both individual component models and state-of-the-art LLMs. **Fig. S2** presents the performance of each individual LLM in our ensemble for ECG labeling (**Fig. S2a**), employment status identification (**Fig. S2b**), and housing status identification (**Fig. S2c**). In all three applications, our ensemble method either matched or outperformed the best individual LLM within the voting committee.

Furthermore, we compared our approach against Meditron-70b[53], a specialized medical LLM, and GPT-4o, the leading commercial LLM at the time of our experiments. As shown in **Fig. S4**, for ECG labeling, our ensemble achieved 98% accuracy compared to 90.5% for Meditron-70b and 97.5% for GPT-4o. **Fig. S4a** shows the confusion matrix for our ensemble method using simple highest vote rule, while **Figs. S4b** and **S4c** present the confusion matrices for Meditron-70b and GPT-4o, respectively. The ensemble approach demonstrated significantly fewer false classifications than either comparison model.

Similarly, **Fig. S5** illustrates that our ensemble method outperformed GPT-4o on SDOH identification for both employment and housing status variables. These results demonstrated that our approach not only competes with state-of-the-art models but can exceed their performance through the collective intelligence of multiple diverse LLMs.

## Hallucination mitigation

A significant advantage of our ensemble approach is its ability to reduce errors caused by LLM hallucinations—false but appearing plausible information generated by language models. The ensemble architecture provides robust protection against hallucinations by requiring multiple independent agents to agree on a classification, substantially reducing the probability of synchronous errors across multiple LLMs. **Supplementary Table 1** presents a set of examples of hallucinations. **Fig. 6** demonstrates how the ensemble voting mechanism effectively mitigated these errors by requiring consensus of multiple LLM agents, significantly reducing false label generations. This advantage highlights the robustness and reliability of the LLM multiagent ensemble method for clinical data labeling tasks.

We investigated hallucinations in the experiments with 7 selected LLMs for AF200 dataset. Hallucinations appear in 4 types: (1) Hallucination with fabricated fact(s); (2) Confusion with uncertainty extent, in which LLMs failed to discern the levels of "probable" and "possible"; (3) Confusion with misunderstandings terms and concepts; (4) Confusion with self-contradiction, in which a LLM generates inconsistent or self-contradiction content. **Table 1** summarizes the frequency of these four types and their distributions.

Table 1 Hallucination types with 7 selected models in AF200 dataset (200 cases)

| Predictor | Hallucination Type | | | | Sum | Percentage in AF200 (200 cases) |
|---|---|---|---|---|---|---|
| | Hallucination with fabricated fact | Confusion with uncertainty extent | Confusion with misunderstand. | Confusion with Self-contradiction | | |
| Beluga (70b) | 5 | 1 | 0 | 0 | 6 | 3.0% |
| Gemma (7b) | 7 | 6 | 0 | 0 | 13 | 6.5% |
| Llama3 (70b) | 0 | 1 | 1 | 1 | 3 | 1.5% |
| Mistral-openorca (7b) | 1 | 4 | 4 | 0 | 9 | 4.5% |
| Openhermes (7b) | 2 | 2 | 2 | 2 | 8 | 4.0% |
| Qwen (72b) | 0 | 2 | 2 | 1 | 5 | 2.5% |
| qwen2 (72b) | 1 | 1 | 1 | 0 | 3 | 1.5% |
| **Sum** | 16 | 17 | 10 | 4 | **47** | |
| **Hallucination Type Distribution** | 34% | 36.2% | 21.3% | 8.5% | | |

Note: For one case, if a LLM has a hallucination error which can be categorized in multiple hallucination types, this table counts only the most significant type.

# Discussion

Our study presents an efficient and effective LLM multiagent ensemble method leveraging diverse LLMs for the purpose of automated labeling of clinical text data, demonstrating significant advantages in accuracy, efficiency, and hallucination reduction. We applied this approach to two substantial healthcare challenges: labeling a large corpus of ECG reports for AF and extracting social determinants of health from clinical notes. Currently, the development of LLMs capabilities for healthcare focus on healthcare foundation models, datasets and benchmarks,[54-58] as well as various novel utilization of a selected specific language model for a specific task in healthcare.[59-67] We focus on exploring a LLM multiagent ensemble approach using a group of diverse LLMs for healthcare text data labeling.

## Summary of clinical data labeling

Our primary application—labeling the MIMIC-IV ECG dataset—addresses a critical need in cardiovascular research. Despite containing 623,566 ECG reports, this important new dataset lacked structured labels for AF, limiting its utility for developing advanced AI models for arrhythmia detection. Through our approach of first filtering AF-relevant cases with logical expressions and then applying the LLM multiagent ensemble method, we achieved over 98% accuracy in classification for AF-relevant cases, while dramatically reducing the resource requirements compared to traditional labeling approaches. Notably, the ensemble method labeled 68,165 AF-relevant ECG reports in approximately two weeks (347 hours) on a single Nvidia A100 GPU node. By comparison, manual labeling by a cardiologist, which we estimated at 200-300 reports per day based on our team's experience, would require approximately 272.66 days (8.8 months) for the 68,165 AF-relevant cases or about 2,494.26 days (6.83 years) for the entire dataset—an impractical timeframe for most research applications. This represents an estimated 99% reduction in labeling time required, transforming a largely knowledge-demanding, time-consuming, labor-intensive, and usually error-prone task into an efficient automatic one while maintaining clinic-grade accuracy. It is worthy to point out that this work exhibits effectiveness of data labeling for the ECG reports that mostly follow AHA/CCC/ACCF guideline[8]. For other data, some specifics (e.g. prompt) need to adapt correspondingly but the principle and methodology are generally applicable.

Our second application demonstrated the generalizability of the LLM multiagent ensemble method in accurately identifying SDOH from clinical notes. The LLM multiagent ensemble method achieved impressive accuracy scores (**Fig. 5**): 95% overall accuracy for employment status labeling, improving to 97.6% with higher voting thresholds, and 99.7% accuracy for housing status, further increasing to 99.9% with stricter thresholds. These outcomes illustrate that our method can reliably extend to various clinical labeling tasks beyond ECG interpretation.

Notably, the ensemble approach consistently outperformed individual LLM models, including specialized medical LLMs (Meditron-70b) and state-of-the-art commercial models (GPT-4o), as shown in **Figs 4, 5, S2, S4&S5**. This performance superiority highlights the inherent advantage of ensemble learning, where aggregating outputs from diverse models mitigates biases, enhances accuracy, and ensures robustness. The demonstrated flexibility and scalability suggest that this LLM multiagent ensemble approach provides a practical and effective solution for diverse clinical annotation challenges.

## Design consideration, rationale, and benefits

For our motivating application of automatic annotation of a large volume of ECG reports, we adapted the framework of LLMs powered multiagent system (LLM-MAS in short). Generally, LLM-MAS is a collection of interacting LLM agents, leveraging collective intelligence. Each LLM agent is a system using a LLM as "brain", sensing the environment, autonomous making decisions, acting on the environment. Those agents communicate and collaborate to solve complex tasks. Each agent may play a different role based on their different "expertise". Depending on the problem, the interactions among agents can be from simple (e.g. voting) to complex (such as teaming, debating, et al.). By this architecture, LLMs are wrapped within each agent with flexible and extendible agentic workflows. An alternative framework is ensemble LLMs, in which each LLM can be regarded as a "learner" or model in an ensemble learning framework. The ensemble method can be simple one like voting or more sophisticated one like stacking. We select LLM-MAS framework, mainly because its advantages of generality and extendibility. That is, LLM-MAS is a general framework for solving a broad range of problems and is flexible and easy to extend with various external functions and tools as well as modules for new problems to solve. If a task needs, ensemble methods can be easily used within LLM-MAS framework. For example, stacking may improve the predication accuracy, by identifying some implicit conditions, where some LLM agents perform better and some others perform worse, and learning to dynamically adapt optimal weights on those LLM agents' outputs. We will apply the method in future projects as needed. More advantages of using this framework will be discussed later in this section.

In our design rationale, in addition to leverage the advantages of emerging technologies, an important consideration is – *to be general, simple and easy to use*. Majority vote, used in both MAS and ensemble method, is general, straightforward, and easy for medical professionals to understand and use. For this rationale, we selected majority vote method. As our results show, it achieved excellent performance in ECG report labeling as well as in the SDOH application.

Practically, this LLM multiagent ensemble architecture offers several specific benefits: (1) Independent clinical reasoning: By treating each LLM agent as an autonomous entity, we simulate independent clinical opinions similar to how multiple specialists might evaluate the same case, each applying their unique knowledge and reasoning patterns. (2) Explicit decision authority: The agent framework transforms LLM outputs from passive suggestions into active clinical judgments that carry decision-making weight within a defined governance structure (the voting mechanism). (3) Assured performance: By utilizing each LLM as the reasoning engine of an agent, this approach allows us to gain the strength of LLM in common knowledge and reasoning, and at the same time to enable controlled, transparent, and manageable application of LLMs, rather than using them directly as end-to-end tools. (4) Accountability and interpretability: LLM multiagent ensemble provides clearer attribution of decisions, enabling analysis of which agents contributed to correct or incorrect classifications and why certain patterns of agreement or disagreement emerge. (5) Modular system architecture: The agent paradigm facilitates a flexible architecture where individual agents can be added, removed, or replaced without redesigning the entire system, allowing for continuous improvement as newer LLMs become available. (6) Domain-specific expertise development: Developing LLMs as agents enables specialized prompt engineering and fine-tuning for specific clinical tasks, allowing each agent to develop domain expertise beyond what a general-purpose model might achieve. (7) Parallel process: with the majority vote method, there is no communication among LLM agents during their data labeling processes, therefore, their data labeling processes can run in parallel. Voting can be conducted after all labeling process completed. This agent-based approach transforms LLMs from text generation tools into active participants in a clinical decision-making process, creating a more robust and interpretable system that better emulates human expert collaboration while maintaining computational efficiency.

Finally, our ensemble method addresses one of the most challenging limitations of LLMs - hallucination.[23 66 68] The generation of plausible but factually incorrect information poses substantial risks in clinical settings, potentially leading to misdiagnosis or inappropriate management decisions. Our LLM multiagent ensemble approach provides robust protection against hallucinations through its voting mechanism, as demonstrated in Fig. 6. The probability of multiple diverse LLMs independently generating identical hallucinations is vanishingly small, creating an effective safeguard that increases confidence in the system's outputs. The configurable voting threshold in our ensemble architecture offers a valuable mechanism for balancing automation efficiency against safety considerations. In high-stakes applications where false negatives carry significant consequences, higher thresholds can be implemented to direct ambiguous cases to expert review. Conversely, in scenarios where rapid processing is prioritized and some margin of error is acceptable, lower thresholds can maximize automation while still maintaining ensemble benefits over single-model approaches.

## Further discussion

Our work has several limitations. First, while we demonstrated significant improvements in hallucination reduction, complete elimination of this phenomenon remains challenging, particularly in cases with complex or ambiguous clinical descriptions. Second, although a careful designed logical filter is an efficient method for a large volume of data like the MIMIC ECG dataset, it is still possible to introduce bias for missing some unregularly expressed cases. Third, as an automatic AF annotation solution for ECG reports using LLM multiagent ensemble method, the effectiveness of this method is validated by a large-scale MIMIC ECG dataset; while additional dataset from a different source would help for better validation. For data labeling in different domain, some specifics (e.g. prompt) need to be adapted correspondingly but the principle and methodology are generally applicable.

Future research directions include exploring more sophisticated interactions among agents (e.g. debate) and adaptive weighting schemes that assign greater influence on better-performing agents for specific task subtypes and investigating the optimal ensemble size and composition for various healthcare applications.

In conclusion, our LLM multiagent ensemble method represents a significant advancement in automated clinical text analysis, offering a robust and scalable solution to address the persistent data labeling bottleneck in healthcare AI while mitigating the risks of hallucinations. By enabling efficient, accurate annotation of large-scale clinical datasets, our work helps bridge the gap between abundant unstructured healthcare data and the structured labels required for developing next-generation clinical AI systems.

# Acknowledgement

This work was partially supported by National Institutes of Health under award numbers P50CA70907, R35GM136375, R01GM140012, R01GM141519, and U01AI169298; the Cancer Prevention and Research Institute of Texas under award numbers RP230330 and RP240521.

**Fig. 1: LLMs multiagent ensemble framework for ECG report annotation and hallucination mitigation**

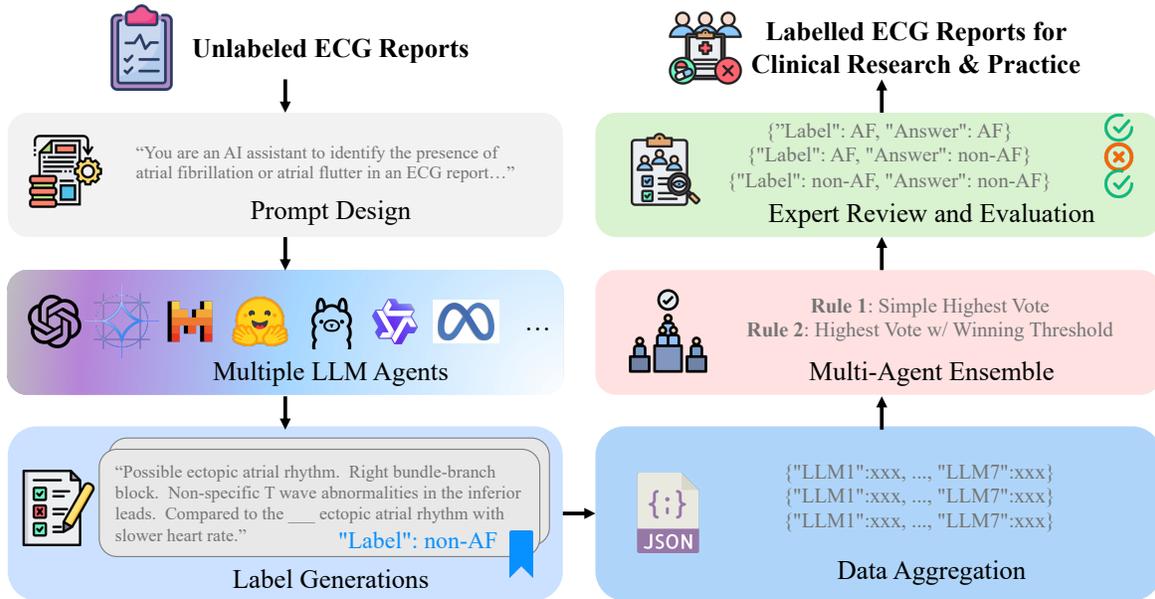

For a large volume of unlabeled MIMIC-IV ECG reports, (1) a prompt is designed for AF-related label generations; (2) a pool of diverse LLMs are selected, as "brain" of each LLM agent; (3) each LLM agent uses the designed prompt to generate labels of "AF", "Non-AF", and "Uncertain" for each ECG report; (4) each ECG report's labels generated by LLM agents are aggregated together; (5) a highest vote with winning threshold mechanism is applied to decide the final label of each ECG report, including label "Review" for expert review when consensus thresholds are not met; (6) human expert(s) manually annotate the ECG reports labeled as "Review"; human expert(s) conduct review and audition for a small selected subset of ECG to ensure the validity of the label generation by LLM multiagent ensemble method. This simple and easy to use mechanism can efficiently conduct automatic annotation of large-scale ECG report data and effectively reduce LLM hallucinations.

**Fig. 2: Data processing workflow of MIMIC-IV ECG data labeling**

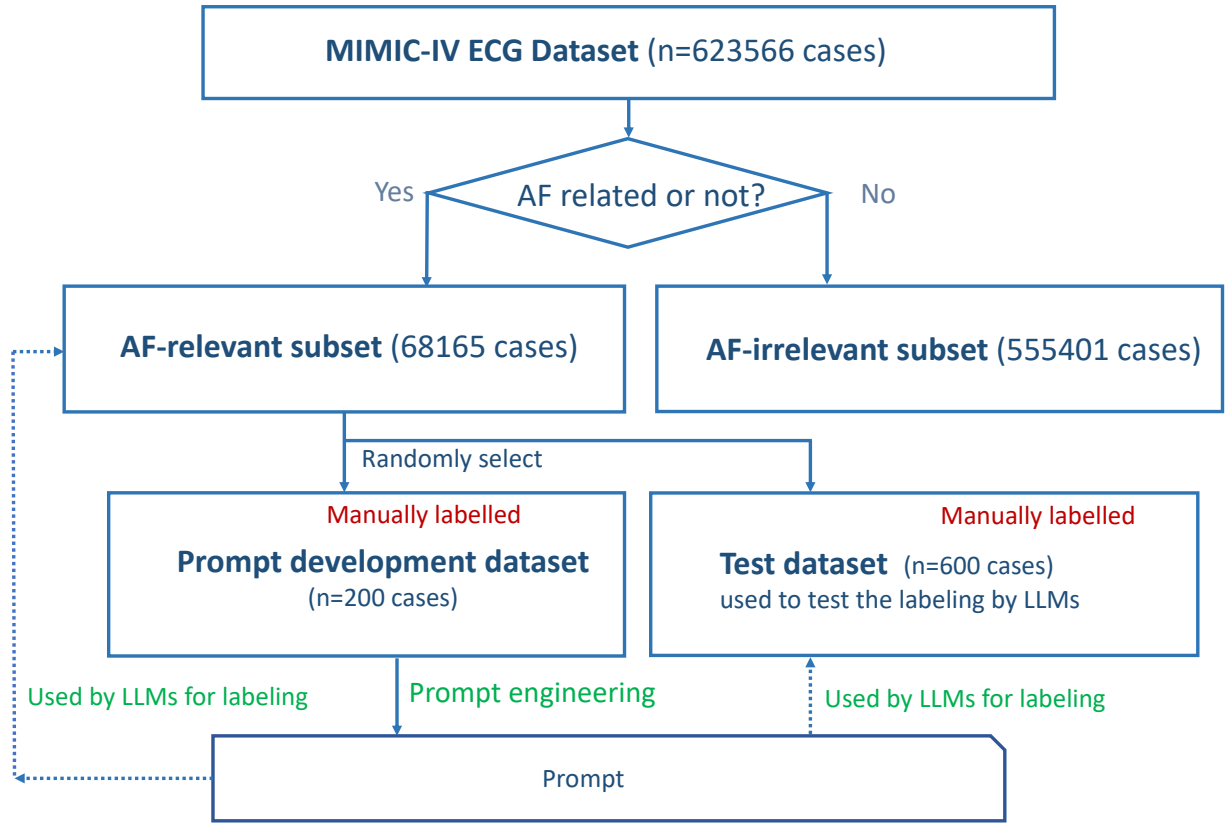

Schematic representation of the data processing workflow for the 623,566 ECG reports in MIMIC-IV. Initial logical filtering identified 68,165 AF-relevant reports (10.93%) containing terminology related to atrial fibrillation or flutter, while 555,401 reports (89.07%) were classified as AF-irrelevant. From the AF-relevant subset, 200 randomly selected cases were manually labeled by cardiologists to develop the classification prompt (AF200 dataset), and an additional 600 cases were manually annotated to create the test dataset for performance evaluation. The LLM multiagent ensemble method was applied to all AF-relevant cases, with verification confirming absence of AF diagnoses in the AF-irrelevant subset.

# Fig. 3: Prompt used for ECG data labeling

**System Message**

You are an AI assistant helping correctly understand the diagnostic statements made by a cardiologist in ECG reports. You will help to identify the presence of atrial fibrillation (AFib) or atrial flutter (AFL) stated by the cardiologist in a given ECG report.

An ECG report consists of a cardiologist's diagnostic statements of cardiac rhythms. An ECG report may begin with the primary rhythm stated with or without specific supporting evidence of rhythm specificity. A certainty modifier (such as 'probable' or 'possible') may be applied to the diagnosed primary rhythm, to express the cardiologist's confidence about his diagnosis. Typically, 'probable' before the diagnosed primary rhythm is used to express high confidence, and 'possible' before the diagnosed primary rhythm is used to express multiple exclusive possibilities of not only the primary rhythm but also some different secondary rhythms considered. Diagnostic statements regarding other heat diseases (such as 'myocardial infarction') may coexist with diagnostic statement regarding AF. These are not diagnostic statements about secondary rhythms.

**Instruction**

You will read an ECG report to be provided at the end. This report is among a series of ECG reports. Each ECG report contains a cardiologist's diagnosis of cardiac rhythms in one specific tracing, called "current tracing" in the report. Your task is to classify the cardiologist's diagnosis for the current tracing into five classes:

1. "AF", if the report surely assesses the presence of atrial fibrillation (AFib) or atrial flutter (AFL) with complete certainty. Let AF be AFib OR AFL. In this class of cases, the estimated probability of AF (AF_pr) shall be 1.0.

2. "Probable AF", if the report assesses AFib or AFL with higher confidence, where there is no secondary rhythm beyond AFib/AFL explicitly stated. In this class of cases, the report describes the primary rhythm as 'probable atrial fibrillation/flutter', 'most likely fibrillation/flutter', 'consistent with fibrillation/flutter', or other equivalent terms. In this class of cases, the estimated AF_pr could be smaller than 1.0 but close to 1.0, depending on the level of confidence shown in the cardiologist's diagnostic statements.

3. "Possible AF", if the report assesses AFib or AFL with lower confidence, where AFib/AFL is among multiple possibilities of different cardiac rhythms. In this class of cases, the report may state as 'possible atrial fibrillation/flutter', "could/may be atrial fibrillation/flutter", "appears to be atrial fibrillation/flutter", "cannot excluded atrial fibrillation/flutter", or other equivalent terms. Based on the cardiologist's diagnostic statements, estimate a probability of AF (denoted as AF_pr); in most uncertain cases, the estimated AF_pr shall be 0.5. Please be aware that 1.0 corresponds to completely certain being AF, 0.50 represents most uncertain, and 0.0 for completely certain being "Not AF".

4. "Not AF", if the report identifies the cardiac rhythm as normal sinus rhythm or other arrhythmia rather than atrial fibrillation/flutter, such as 'atrial tachycardia', 'ventricular tachycardia', 'sinus bradycardia', and many others. In this class of cases, the estimated AF_pr shall be 0.

5. "Not Specified", if no rhythm is directly described for the current tracing; instead, the report just simply refers to the previous tracing or report.

Provide your answer in JSON data as follows.

{

"Diagnosis": <"AF"/"Probable AF"/"Possible AF"/"Not AF"/"Not Specified">,

"AF_pr": <Estimated probability of the presence of AFib or AFL in the current tracing, as a number between 0 and 1 to two decimal places (e.g. 0.85)>,

"Explanation": <Explanation of your answers with support of the evidence presented in the ECG report>

}

Please do remember that "probable atrial fibrillation (or atrial flutter)" is a case of "Probable AF" and it is NOT "Possible AF"; similarly, "possible atrial fibrillation (or atrial flutter)" is a case of "Possible AF" and it is NOT "Probable AF".

Please output in a single JSON file and do not output anything else. Make sure the output JSON data is in legitimate JSON format. In your JSON output, ensure each value is embraced with a pair of straight double quotes; do not use any double quotation marks within the string of a value.

**ECG Report**:

The prompt we used with 7 selected LLMs for LLM multiagent ensemble labeling of ECG reports.

**Fig. 4: Evaluation of LLM multiagent ensemble method on test dataset**

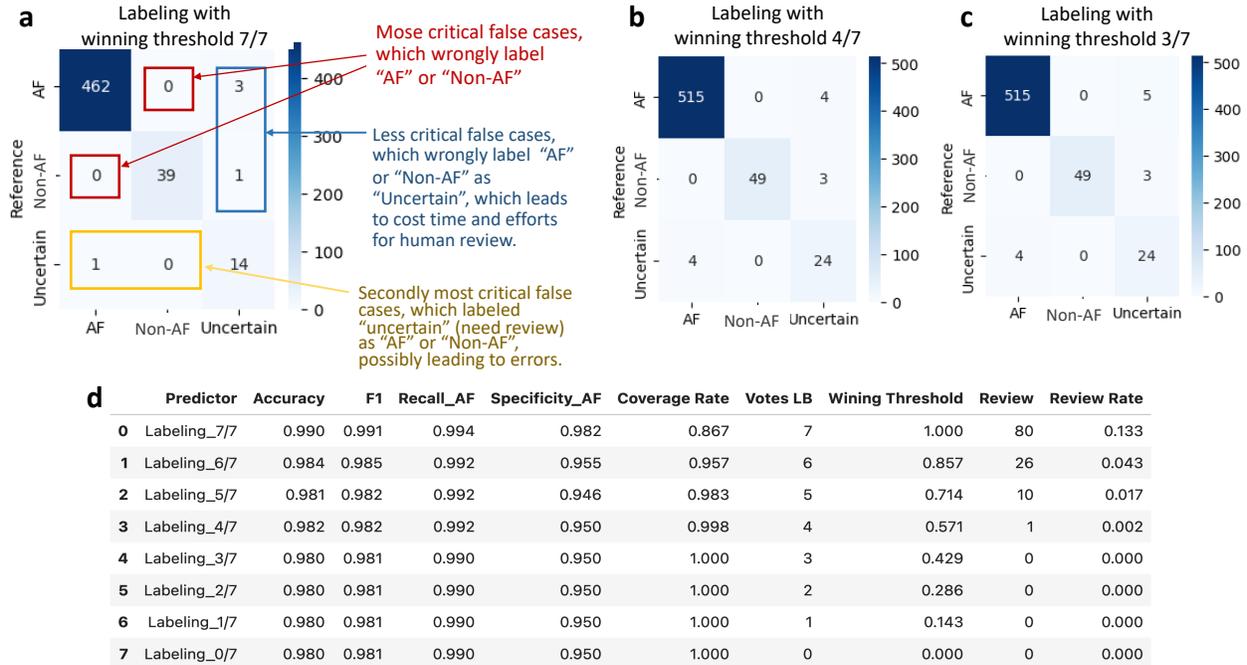

a. Confusion matrix of labeling with winning threshold (MinVoteRate) 7/7 (100%). b. Confusion matrix of labeling with winning threshold 4/7 (at least 50%). c. Confusion matrix of labeling with winning threshold 3/7 (the same for 3/7 ~ 0/7). d. The performance metrics of labeling with different winning thresholds. Recall and Specificity are for AF. With the increase of winning threshold, the accuracy increases; however, the number of cases need review also increase significantly, particularly in high thresholds close to 100%.

**Fig. 5: Evaluation of LLM multiagent ensemble method on identifying SDOH**

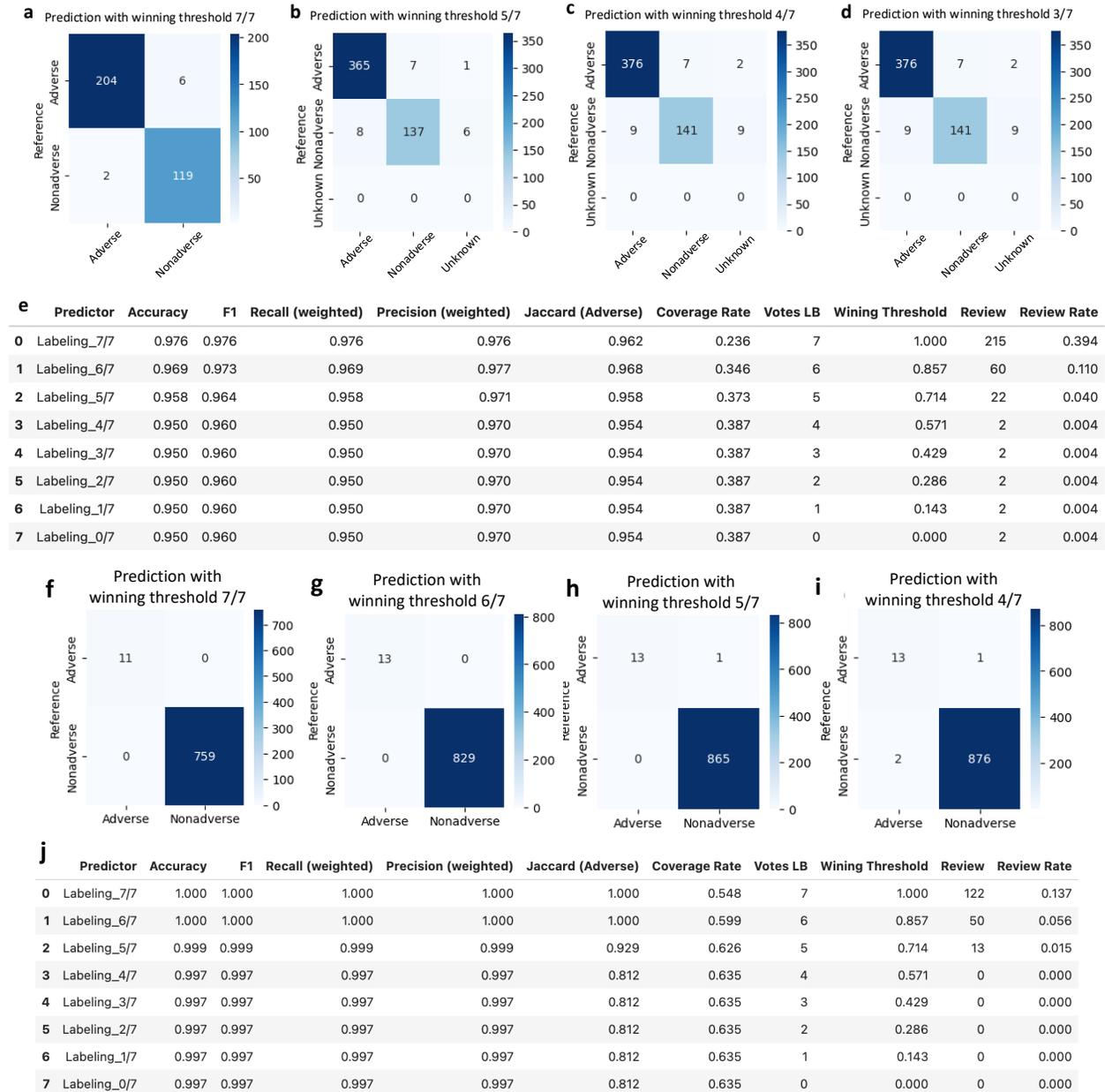

The evaluation of LLM multiagent ensemble method applied to identify social determinants of health with a randomly selected 20% of MIMIC-SBOH data on *employment* status (a to e) and housing status (f to j). Jaccard index is defined as TP/(TP+FP+FN). **a**. Confusion matrix by applying winning threshold of 7/7 (100%). **b.** Confusion matrix by applying winning threshold of 5/7. **c**. Confusion matrix by applying winning threshold of 4/7. **d**. Confusion matrix by applying winning threshold of 3/7 (2/7 ~ 0/7 have the same confusion matrix.). **e**. Performance metrics with different winning thresholds. Similar to ECG labeling, with the increase of winning threshold, the accuracy increases; however, the number of cases need review also increase significantly. **f**. Confusion matrix by applying winning threshold of 7/7 (100%). **g.** Confusion matrix by applying winning threshold of 6/7. **h**. Confusion matrix by applying winning threshold of 5/7. **i**. Confusion matrix by applying winning threshold of 4/7 (3/7 ~ 0/7 have the same confusion matrix.). **j**. Performance metrics with different winning thresholds. Similar to ECG labeling, with the increase of winning threshold, the accuracy increases; however, the number of cases need review also increase.

**Fig. 6: LLM multiagent ensemble method reduces hallucination errors**

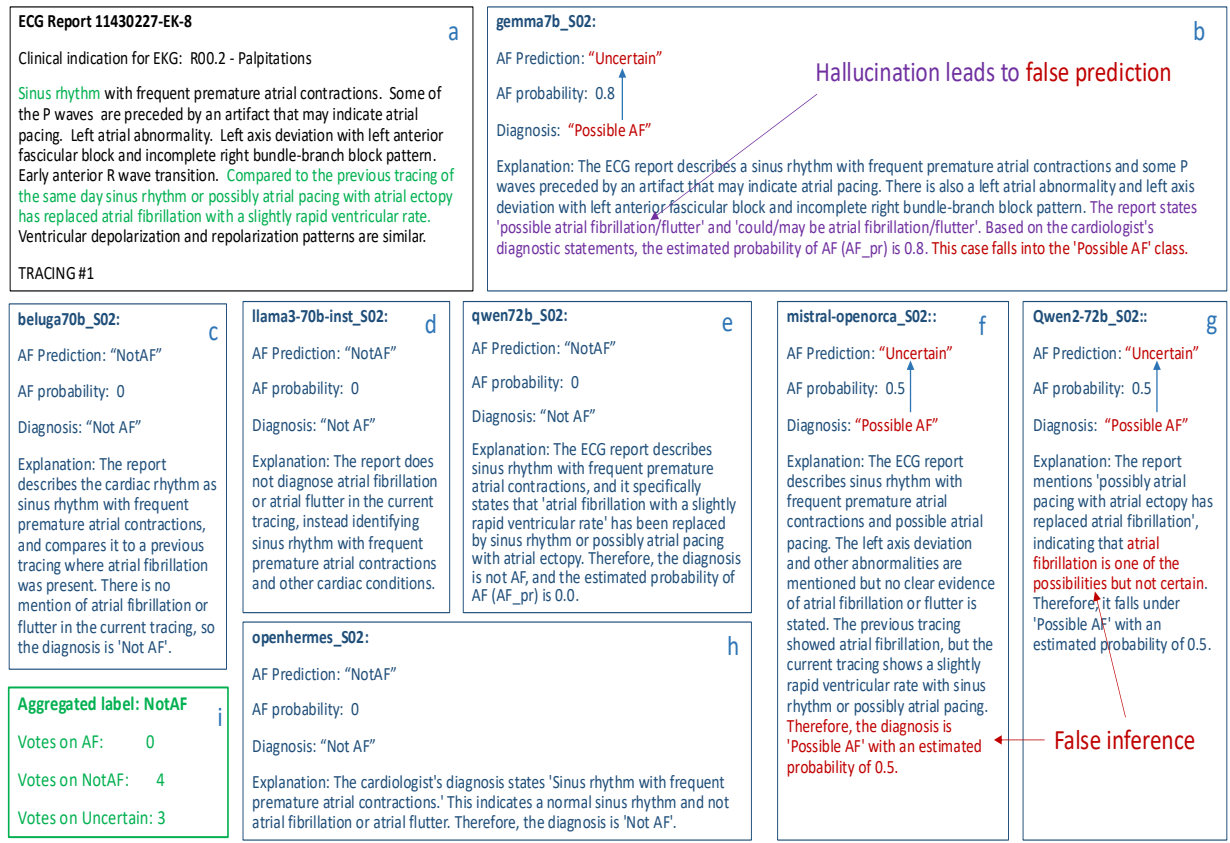

The LLM multiagent ensemble method reduces the chance of errors caused by LLM hallucinations. Case 11430227-EK-8: **a.** ECG report; **b.** prediction by gemma (7b) with prompt S02, noted as gemma7b_S02; **c.** prediction by beluga (70b); **d.** prediction of llama3 (70b-instruct); **e.** prediction of qwen (72b); **f.** prediction of mistral-openorca (7b); **g.** prediction of qwen2 (72b); **h.** prediction of openhermes (7b); **i.** the LLM multiagent ensemble labeling of the case. All the above LLMs use the same prompt S02. Each predictor is noted as <name of LLM>_<prompt id>, e.g., gemma7b_S02. The ECG report shows a non-AF case in current tracing. Gemma7b_S02 hallucinated "Possible AF" with fabricated facts that "The report states 'possible atrial fibrillation/flutter' and 'could/may be atrial fibrillation/flutter'", which is not true in the original ECG report. This hallucination led the LLM to predict "Possible AF" with probability of 0.8, thus making false prediction. Mistral-openorca_S02 and Qwen2-72b_S02 did not hallucinate but made false inference. Four LLM agents, beluga70b_S02, llama3-70b-instruct_S02, openhermes_S02, and qwen72b_S02, made correct prediction as "Non-AF". By the ensemble rule, the final ensemble label is "Non-AF" which is consistent with the ECG report.

# Supplementary information

## Fig. S1: Performance of 7 selected LLMs in AF200 dataset

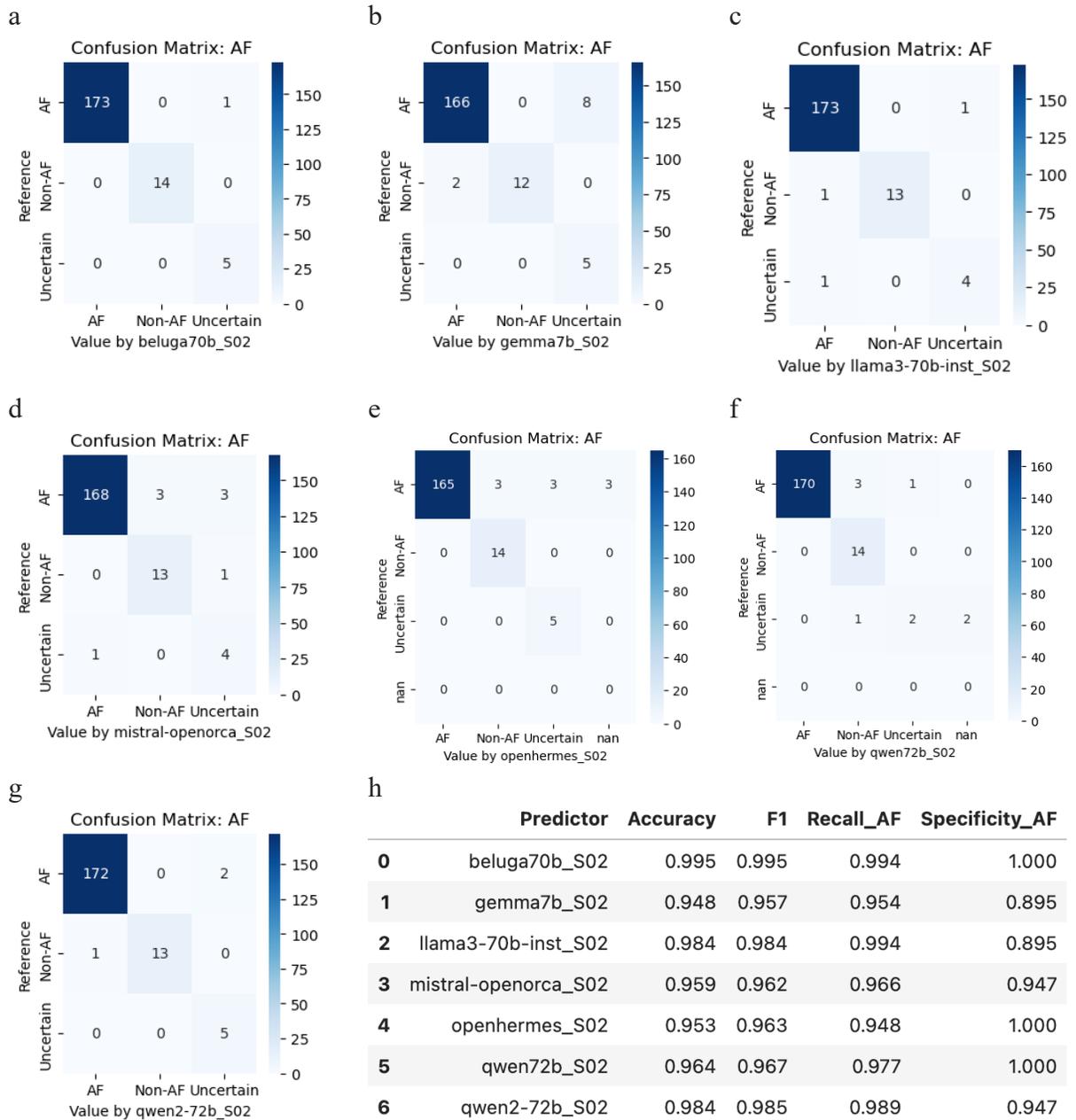

Label "nan" represents the cases where a LLM failed to respond with requested output. "S02" is the code for the prompt illustrated in Fig. 2.

**Fig. S2: 7 LLMs used in our experiments and their performance over test dataset**

**a**

|   | Predictor | Accuracy | F1 | Recall_AF | Specificity_AF |
|---|---|---|---|---|---|
| 0 | beluga70b_S02 | 0.973 | 0.974 | 0.988 | 0.925 |
| 1 | gemma7b_S02 | 0.925 | 0.934 | 0.954 | 0.850 |
| 2 | llama3-70b-inst_S02 | 0.968 | 0.970 | 0.983 | 0.938 |
| 3 | mistral-openorca_S02 | 0.957 | 0.959 | 0.975 | 0.912 |
| 4 | openhermes_S02 | 0.933 | 0.945 | 0.950 | 0.925 |
| 5 | qwen72b_S02 | 0.938 | 0.945 | 0.956 | 0.988 |
| 6 | qwen2-72b_S02 | 0.978 | 0.979 | 0.988 | 0.962 |

**b**

|   | Predictor | Accuracy | F1 | Recall (weighted) | Precision (weighted) | Jaccard (Adverse) |
|---|---|---|---|---|---|---|
| 0 | beluga70b_P03 | 0.927 | 0.945 | 0.927 | 0.963 | 0.929 |
| 1 | gemma7b_P03 | 0.676 | 0.688 | 0.676 | 0.833 | 0.548 |
| 2 | llama3-70b-inst_P03 | 0.929 | 0.945 | 0.929 | 0.964 | 0.932 |
| 3 | mistral-openorca_P03 | 0.886 | 0.908 | 0.886 | 0.932 | 0.876 |
| 4 | openhermes_P03 | 0.868 | 0.903 | 0.868 | 0.946 | 0.879 |
| 5 | qwen72b_P03 | 0.930 | 0.942 | 0.930 | 0.955 | 0.925 |
| 6 | qwen2-72b_P03 | 0.927 | 0.944 | 0.927 | 0.964 | 0.935 |

**c**

|   | Predictor | Accuracy | F1 | Recall (weighted) | Precision (weighted) | Jaccard (Adverse) |
|---|---|---|---|---|---|---|
| 0 | beluga70b_P03 | 0.978 | 0.986 | 0.978 | 0.994 | 0.684 |
| 1 | gemma7b_P03 | 0.998 | 0.998 | 0.998 | 0.998 | 0.857 |
| 2 | llama3-70b-inst_P03 | 0.981 | 0.990 | 0.981 | 0.999 | 0.857 |
| 3 | mistral-openorca_P03 | 0.979 | 0.987 | 0.979 | 0.995 | 0.722 |
| 4 | openhermes_P03 | 0.902 | 0.937 | 0.902 | 0.985 | 0.151 |
| 5 | qwen72b_P03 | 0.920 | 0.948 | 0.920 | 0.987 | 0.197 |
| 6 | qwen2-72b_P03 | 0.934 | 0.956 | 0.934 | 0.987 | 0.224 |

The seven LLMs we used for the LLM multiagent ensemble method experiments and their performance. Jaccard index is defined as TP/(TP+FP+FN). **a.** ECG data labeling performance of 7 individual LLMs, evaluated with test data 1. "S02" is the name of the used prompt. **b.** SDOH employment status identification performance of 7 individual LLMs. **c.** SDOH housing status identification performance of 7 individual LLMs. "P03" is the name of the used prompt.

## Fig. S3: The labeling result by LLM multiagent ensemble method

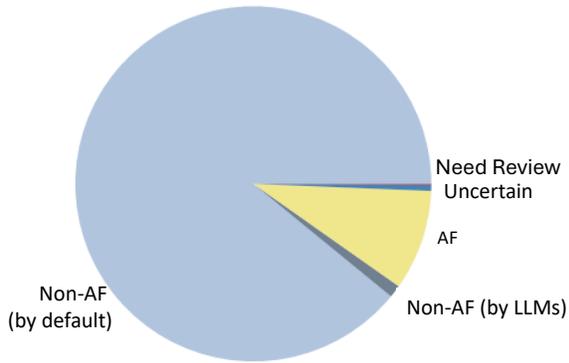

| Labels | Numbers | Percentage |
|---|---|---|
| **AF-irrelevant cases** | **555401** | **89.07%** |
| Non-AF (by default) | 555401 | 89.07% |
| | | |
| **AF-relevant cases (labeled by LLM agents)** | **68165** | **10.93%** |
| Non-AF (by LLMs) | 7141 | 1.15% |
| AF | 57029 | 9.15% |
| Uncertain | 3635 | 0.58% |
| Need Review | 339 | 0.05% |
| | | |
| **Total number of cases** | **623566** | **100.00%** |

Labels include: "AF" (either "atrial fibrillation" or "atrial flutter"), "Non-AF" (neither "atrial fibrillation" nor "atrial flutter"), "Non-AF" (labeled by LLMs), "Uncertain" (no sufficient information to tell it is AF or non-AF), and "Review" (need human experts to review, due to high disagreement among the votes from LLM agents). Cases with either "Uncertain" or "Review" need to be reviewed (but for different reasons). The final ensemble label distribution over the whole ECG dataset. All 555401 AF-irrelevant cases were labeled as "Non-AF".

## Fig. S4: Performance comparison on test dataset: the LLM multiagent ensemble vs Meditron 70b and GPT-4o

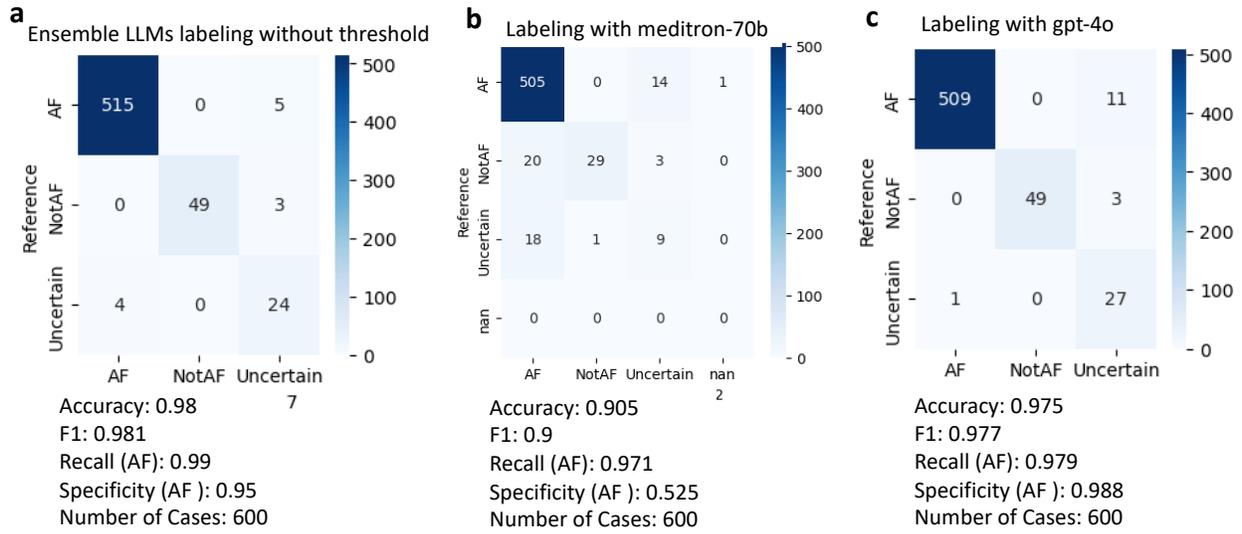

**a.** Ensemble LLMs labeling without threshold
Accuracy: 0.98
F1: 0.981
Recall (AF): 0.99
Specificity (AF ): 0.95
Number of Cases: 600

**b.** Labeling with meditron-70b
Accuracy: 0.905
F1: 0.9
Recall (AF): 0.971
Specificity (AF ): 0.525
Number of Cases: 600

**c.** Labeling with gpt-4o
Accuracy: 0.975
F1: 0.977
Recall (AF): 0.979
Specificity (AF ): 0.988
Number of Cases: 600

The labeling performance comparison between the LLM multiagent ensemble, Meditron-70b (a medical LLM) and GPT-4o, evaluated on ECG test dataset. **a.** performance of the ensemble labeling with 7 open source LLMs using simple highest vote rule (without winning threshold); **b.** performance of meditron-70b; **c.** performance of GPT-4o. When a winning threshold is applied, the LLM multiagent ensemble method has improved performance at the cost of increasing the number of cases need to be reviewed. Please refer to Fig. 4d for details.

## Fig. S5: Performance comparison: GPT-4o vs LLM multiagent ensemble on identifying social determinants of health

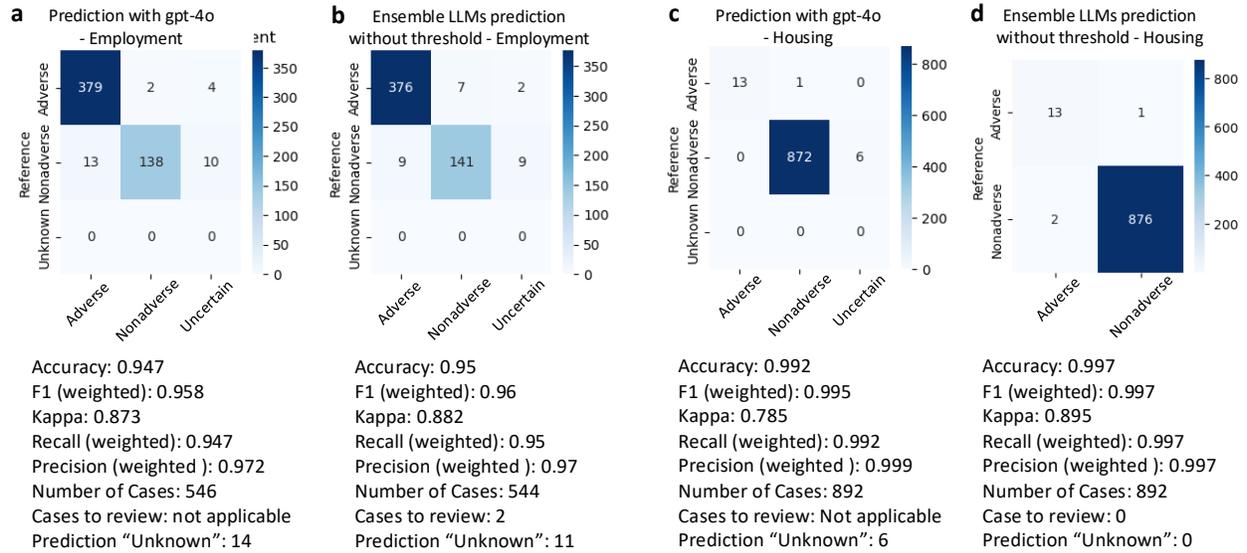

Accuracy: 0.947
F1 (weighted): 0.958
Kappa: 0.873
Recall (weighted): 0.947
Precision (weighted ): 0.972
Number of Cases: 546
Cases to review: not applicable
Prediction "Unknown": 14

Accuracy: 0.95
F1 (weighted): 0.96
Kappa: 0.882
Recall (weighted): 0.95
Precision (weighted ): 0.97
Number of Cases: 544
Cases to review: 2
Prediction "Unknown": 11

Accuracy: 0.992
F1 (weighted): 0.995
Kappa: 0.785
Recall (weighted): 0.992
Precision (weighted ): 0.999
Number of Cases: 892
Cases to review: Not applicable
Prediction "Unknown": 6

Accuracy: 0.997
F1 (weighted): 0.997
Kappa: 0.895
Recall (weighted): 0.997
Precision (weighted ): 0.997
Number of Cases: 892
Case to review: 0
Prediction "Unknown": 0

The performance comparison between GPT-4o and the LLM multiagent ensemble method on identifying social determinants of health. **a.** performance of GPT-4o on employment status identification; **b.** performance of the LLM multiagent ensemble method on employment status identification, using 7 open source LLMs and using highest vote rule (without winning threshold); **c.** performance of GPT-4o on housing status identification; **d.** performance of the ensemble method on housing status identification, using 7 open source LLMs and using highest vote rule (without winning threshold).

**Fig. S6: The architecture, workflow, and data flow of LLM multiagent ensemble application**

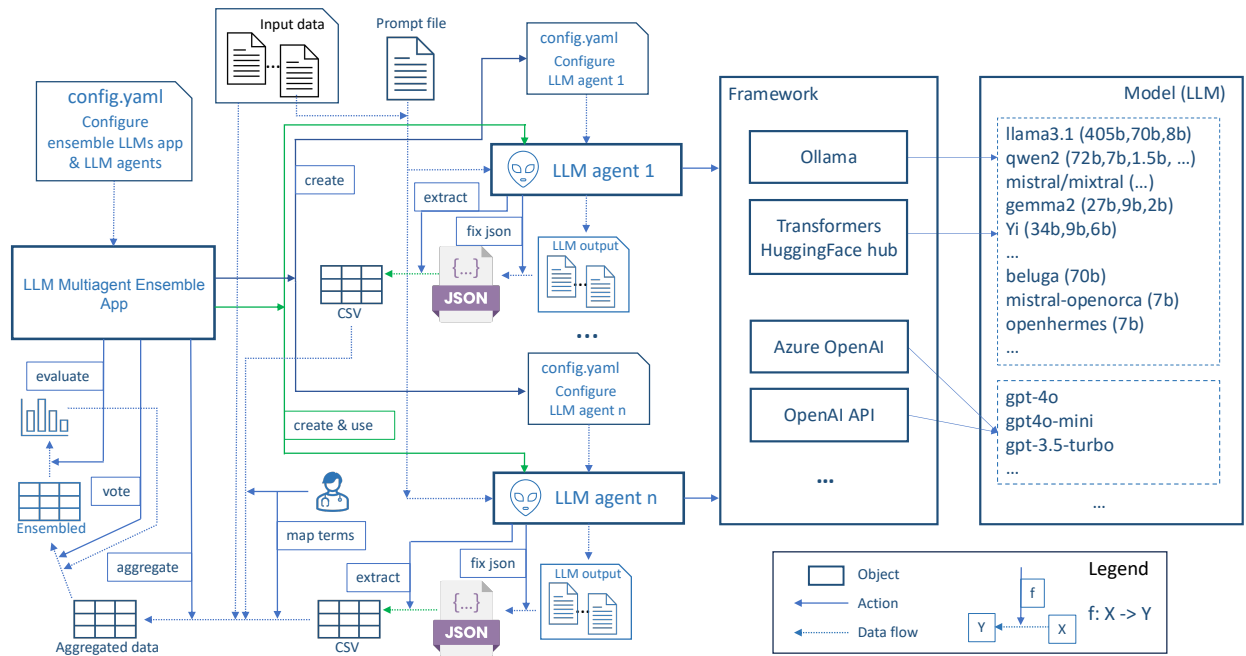

The architecture, workflow, and data flow of the LLM multiagent ensemble application: Configuration yaml file specifies the LLMs to use, the framework to use, the input data path, the prompt file path, and others. LLM multiagent ensemble application object will create a group of specified LLM agents, each of which uses the specified framework to load and run a requested large language model. Each LLM agent will receive the same prompt and the same input data then produce JSON style text outputs. The LLM agent can fix common JSON grammar errors in the LLM output, extract structured data from the JSON data, and save to a CSV file. Human users may provide some term mappings. The LLM multiagent ensemble application aggregates the CSV files from every LLM agent together, uses ensemble rules to vote and calculate the ensembled result, and finally evaluate the ensembled result. The ensemble application then aggregates individual predictions, applies voting rules, calculates consensus labels, and evaluates performance. Optional human-provided term mappings can enhance domain-specific interpretation. This modular architecture enables flexible deployment across diverse clinical text analysis tasks.

**Supplementary Table 1 Examples of hallucinations**

| ECG Report | LLM Generated Contents | Explanation |
|---|---|---|
| **Report 10123949-EK-104:** Clinical indication for EKG: I48.91 - Unspecified atrial fibrillation<br><br>Please see final interpretation under scanned ECG.<br><br>**Ground truth: Not specified**<br><br>(Note: The first line of text in blue indicates the reason for the ECG test and is NOT the description or conclusion about current ECG trial; thus removed in preprocessing) | **Beluga70b: Diagnosis: Probable AF**<br>The report states the presence of atrial fibrillation (or atrial flutter) with complete certainty, indicating a diagnosis of 'AF' and an estimated probability of AF (AF_pr) of 1.0. | Type: Hallucination with fabricated fact<br><br>- Fabricated fact is in red text |
| | **Gemma7b: Diagnosis: Possible AF**<br>The ECG report states 'possible atrial fibrillation/flutter' and describes the rhythm as 'irregular with frequent premature ventricular contractions'. The presence of frequent premature ventricular contractions is suggestive of atrial fibrillation/flutter, however, the report does not provide sufficient evidence to definitively diagnose AFib or AFL. Therefore, the diagnosis is classified as 'Possible AF' with an estimated AF_pr of 0.6. | Type: Hallucination with fabricated facts<br><br>(Fabricated facts are in red text) |
| | **Openhermes: Diagnosis: Possible AF**<br>The cardiologist's diagnostic statement in the ECG report is 'possible atrial fibrillation/flutter'. This indicates a lower confidence in assessing the presence of atrial fibrillation or atrial flutter, as it is among multiple possibilities of different cardiac rhythms. Therefore, this case falls under the category of 'Possible AF', and the estimated probability of AF (AF_pr) is 0.5. | Type: Hallucination with fabricated facts<br><br>(Fabricated facts are in red text) |
| **Report 10115182-EK-50:** Ectopic atrial rhythm at a normal rate. The Q-T interval is prolonged. There are Q waves in the anterior leads consistent with probable myocardial infarction. Diffuse marked ST-T wave changes. Compared to the ___ ventricular pacing is no longer present and **ectopic atrial** | **Gemma7b: Diagnosis: Probable AF**<br>The cardiologist's diagnosis for the current tracing is 'probable atrial fibrillation (or atrial flutter)'. The presence of AFib or AFL is highly probable based on the presence of diffuse marked ST-T wave changes and the absence of ventricular pacing. However, there is no secondary rhythm explicitly stated in the report, therefore the estimated probability of AF (AF_pr) is slightly less than 1.0. | Hallucination with fabricated fact (in red text) |
| **rhythm has replaced atrial flutter.**<br><br>**Ground truth: Non-AF** | **Llama3-70b: Diagnosis: Probable AF**<br>The diagnosis is classified as 'Probable AF' because the cardiologist states that atrial flutter was present in the previous report, but now it has been replaced by an ectopic atrial rhythm. Although the current tracing does not directly describe atrial fibrillation or flutter, the presence of ectopic atrial rhythm and the replacement of atrial flutter suggest a high probability of AFib/AFL | Confusion – Self-contradiction (in red text) |
| | **Qwen2-72b: Diagnosis: Probable AF**<br>The ECG report mentions 'ectopic atrial rhythm' which replaced 'atrial flutter'. This indicates a higher confidence in the diagnosis of an atrial arrhythmia, specifically mentioning atrial flutter as a previous condition. Therefore, it is classified as 'Probable AF' with an estimated probability of 0.90. | Hallucination with fabricated rule (in red text) |
| **Report 10098993-EK-127:**<br>Ventricular paced rhythm. Underlying atrial fibrillation/flutter. Compared to the ___ there is no change.<br><br>**Ground truth: AF** | **Mistral-openorca: Diagnosis: Non-AF**<br>The ECG report states a ventricular paced rhythm with an underlying atrial fibrillation/flutter, but the primary diagnosis is not AFib or AFL. The report compares it to the previous report and mentions no change, which indicates that the current tracing's rhythm is not AFib or AFL. | Type:<br>1. Confusion with misunderstanding<br>2. Confusion with Self-contradiction<br>3. Hallucination with fabricated fact<br><br>- Misunderstanding concept: "paced rhythm" is not an arrhythmia.<br>- Fabricated fact is in red text<br>- Text in red contradicts with text in blue |
| | **Openhermes: Diagnosis: Non-AF**<br>The cardiologist's diagnosis states 'Ventricular paced rhythm. Underlying atrial fibrillation/flutter.' This indicates that the primary rhythm is ventricular paced rhythm, and there is an underlying atrial fibrillation or flutter. Since the primary rhythm is not AFib or AFL, this case falls under 'Not AF' category. | Type:<br>Confusion with misunderstanding<br><br>- Misundersting concept: "paced rhythm" is not an arrhythmia. |
| **Report 10459005-EK-56:** | **Gemma7b: Diagnosis: Possible AF** | Type: |
| *Atrial fibrillation* with a rapid ventricular response. *Possible* prior anterior infarction, age undetermined. Diffuse ST-T wave changes. *Cannot exclude ischemia*. Compared to the previous tracing the rate is increased. Findings are otherwise similar. TRACING #1<br><br>**Ground truth: Probable AF** | The ECG report states 'atrial fibrillation with a rapid ventricular response' and 'cannot exclude ischemia'. The presence of diffuse ST-T wave changes and the inability to exclude ischemia suggest a possible diagnosis of atrial fibrillation. However, the report also states that the findings are similar to the previous tracing, which may indicate that the diagnosis is uncertain. Overall, <span style="color:red">the cardiologist's diagnostic statement is more suggestive of possible atrial fibrillation rather than probable atrial fibrillation.</span> | Confusion with uncertainty judgment<br><br>- Confusion in judging "probable" or "possible" |